%% file: main.tex
\documentclass[10pt,twocolumn,letterpaper]{article}

\usepackage{cvpr}              

\usepackage[pagebackref,breaklinks,colorlinks,allcolors=cvprblue]{hyperref}

\input{preamble}

\definecolor{cvprblue}{rgb}{0.21,0.49,0.74}

\def\paperID{31899} 
\def\confName{CVPR}
\def\confYear{2026}

\title{PAS: A Training-Free Stabilizer for Temporal Encoding in Video LLMs}

\author{
Bowen Sun$^{1,5}$ \quad
Yujun Cai$^{2,3}$\thanks{Corresponding author.} \quad
Ming-Hsuan Yang$^{1,4}$ \quad
Hang Wu$^{1}$ \quad
Yiwei Wang$^{1}$ \\
$^{1}$University of California, Merced \quad
$^{2}$The University of Queensland  \\
$^{3}$Ant Group  \quad
$^{4}$Google DeepMind \quad
$^{5}$Tsinghua University
\\
{\tt\small bowensun2@ucmerced.edu} \\
\href{https://github.com/Bowen-Sun-0728/PAS}{\textcolor{magenta}{\tt \small https://github.com/Bowen-Sun-0728/PAS}} \\
}

\begin{document}
\maketitle
\input{sec/0_abstract}

\input{sec/1_intro}

\input{sec/2_related_work}
\input{sec/4_methodology}

\input{sec/5_experiment}
\input{sec/6_conclusion}
\input{sec/7_acknowledgements}
{
    \small
    \bibliographystyle{ieeenat_fullname}
    \bibliography{main}
}


\end{document}

%% file: preamble.tex
\usepackage{amsmath,amssymb,mathtools}
\usepackage{amsthm}
\usepackage{bm}
\usepackage{algorithm}
\usepackage{algpseudocode}
\usepackage[table]{xcolor}
\usepackage{booktabs}
\usepackage{makecell}   
\usepackage{multirow}
\usepackage{colortbl}
\usepackage{graphicx}
\usepackage{threeparttable}
\usepackage{tikz}
\usetikzlibrary{arrows.meta,patterns}
\usepackage{xcolor}

\newtheorem{theorem}{Theorem}

\newcommand{\phaseop}[1]{\Gamma_{#1}}  

\usepackage[nameinlink,capitalize,noabbrev]{cleveref}

\Crefname{theorem}{Theorem}{Theorems}
\Crefname{lemma}{Lemma}{Lemmas}
\Crefname{proposition}{Proposition}{Propositions}
\Crefname{corollary}{Corollary}{Corollaries}
\Crefname{assumption}{Assumption}{Assumptions}

\Crefname{algorithm}{Algorithm}{Algorithms}
\Crefname{section}{Section}{Sections}
\Crefname{subsection}{Section}{Sections}
\Crefname{equation}{Equation}{Equations}
\Crefname{figure}{Figure}{Figures}
\Crefname{table}{Table}{Tables}

%% file: sec/0_abstract.tex
\begin{abstract}
Video LLMs suffer from temporal inconsistency: small shifts in frame timing can flip attention and suppress relevant frames. We trace this instability to the common extension of Rotary Position Embeddings to video through multimodal RoPE. The induced inverse Fourier time kernel exhibits frame-scale ripples that multiply adjacent frames by different factors, which perturbs attention that should otherwise be governed by the raw query key inner product. We present \textbf{Phase Aggregated Smoothing (PAS)}, a simple, training-free mechanism that applies small opposed phase offsets across heads and then aggregates their outputs. PAS preserves the per-head spectrum magnitude, while the aggregation effectively smooths the temporal kernel and reduces phase sensitivity without changing the positional encoding structure. Our analysis shows that the RoPE rotated logit can be approximated as a content dot product scaled by a time kernel; smoothing this kernel yields Lipschitz stability of attention to small temporal shifts; multi-phase averaging attenuates high frequency ripples while preserving per-head spectra under Nyquist-valid sampling. Experiments on multiple video understanding benchmarks under matched token budgets show consistent improvements with negligible computational overhead. PAS provides a plug-and-play upgrade for robust temporal encoding in Video LLMs.
\end{abstract}

%% file: sec/1_intro.tex
\section{Introduction}

Large multimodal language models extend language-centric architectures to video, enabling open-ended description and reasoning over long temporal contexts. A widely used positional encoding is Rotary Position Embeddings (RoPE), originally designed for discrete text tokens. In Video LLMs, RoPE is generalized to a multimodal variant (M-RoPE) that encodes temporal ($T$), height ($H$), and width ($W$) dimensions. Along the temporal axis, this construction amounts to assigning a finite set of temporal frequencies; each frequency line rotates query and key pairs and induces an independent, periodic contribution to the attention logit. In practice, downstream accuracy is sensitive to sampling details: small changes of frame rate or sampling offsets can alter which motion cues are aggregated per token, exposing fragility in temporal encoding. One of the reasons is that when this text-oriented mechanism is applied to continuous time, the attention distribution may be disrupted.

\begin{figure}[t]
  \centering
  \includegraphics[width=1.0\linewidth, clip, trim=262 570 0 62]{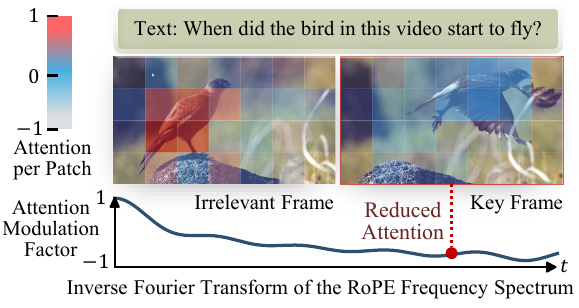}
  \caption{A failure from a real clip~\cite{yt:CJHP6dPjuGY}. Temporal RoPE imposes an interval-dependent gain on attention; when a key frame lies in a low-gain trough of the modulation, it is down weighted compared to less relevant frames, which propagates to downstream errors.}
  \label{fig:intro-attn-instability}
\end{figure}

\begin{figure*}[t]
  \centering
  \includegraphics[width=1.0\linewidth, clip, trim=0 497 742 0]{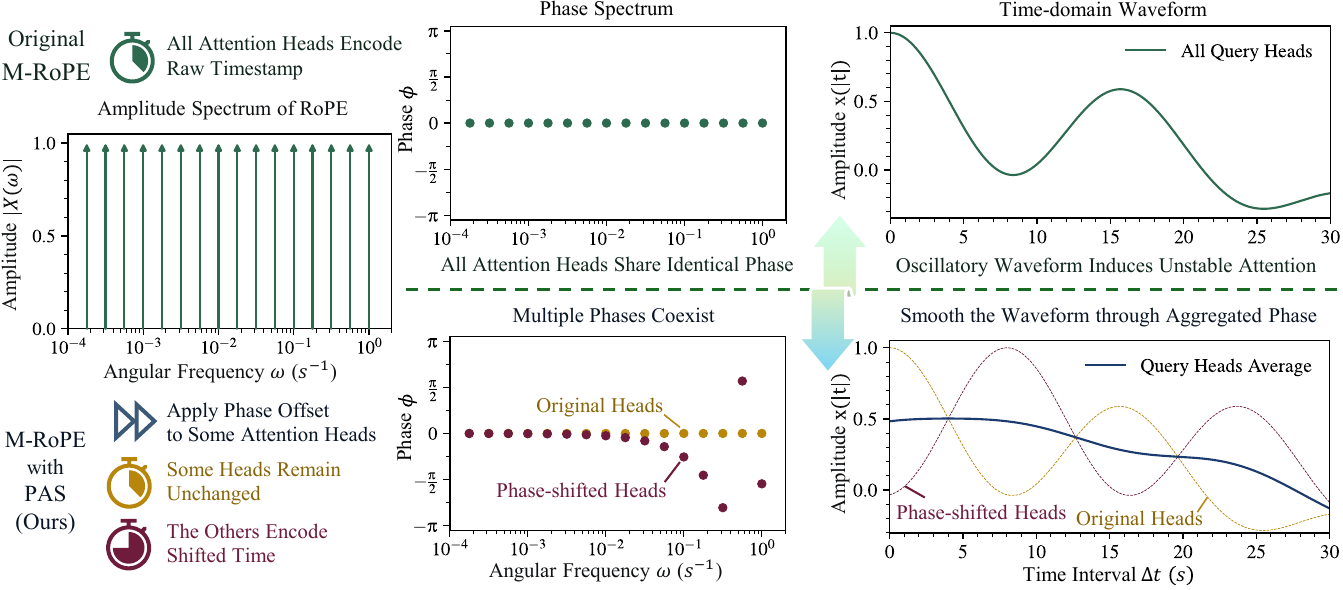}
  \caption{Temporal non-smoothness of the time domain modulation from original M-RoPE (upper) and how Phase Aggregated Smoothing (PAS) mitigates it (lower). PAS assigns small, opposed phase shifts to the query stream per head. Each head preserves its spectrum magnitude because a time shift only rotates phases. Head aggregation then acts as a controlled moving average in time, producing a smoother effective modulation across adjacent frames and reducing low gain induced suppression of key frames.}
  \label{fig:intro-figure}
\end{figure*}

We adopt a Fourier view to address this issue. Let $\{\omega_k\}$ denote the temporal line set and define $m(t)$ as the inverse Fourier time-domain kernel associated with these lines. Under a standard high dimensional approximation, the RoPE rotated logit at relative lag $t$ can be written as the unrotated content dot product multiplied by $\mathrm{Re}\{m(t)\}$; we later formalize this phase modulation view as \cref{thm:1}. Because $m(t)$ is an average of cosines over $\{\omega_k\}$, it is intrinsically rippled at the frame scale, which renders the overall modulation non-smooth. Such non-smoothness means that the attention assigned to frames varies with their relative lag: within specific temporal intervals, the modulation can either amplify or attenuate attention. A concrete failure occurs when a key frame lands in a low gain interval of $m(t)$, causing its evidence to be suppressed or even ignored (\cref{fig:intro-attn-instability}). The upper panel of \cref{fig:intro-figure} visualizes how this interval-dependent modulation undermines temporal stability by imposing gain swings across adjacent frames.

Guided by the idea of a moving average filter, we use multi-head aggregation as a controlled averaging of slightly shifted views of the same temporal kernel, attenuating the high frequency ripple of $m(t)$ while retaining its low frequency trend. We therefore introduce Phase Aggregated Smoothing (PAS), a training-free mechanism that assigns small, opposed phase offsets to attention heads while keeping tokenization and parameters unchanged. All heads operate on the same merged frame tokens; by applying a head specific temporal offset to queries only, each head observes a time shifted version of $m(t)$. Because a time shift changes only phase in the frequency domain, the per-head spectrum magnitude of M-RoPE is preserved. Aggregating these complementary heads yields a locally smoother effective modulation across adjacent frames, reducing variance in the multiplicative factor and stabilizing temporal behavior, while leaving tokens and positional structure unchanged.

We support this design with three core results that preview the analysis to come. First, the phase modulation view implies a Lipschitz type stability principle: when the IFT kernel is smoother, the RoPE rotated logit changes at most linearly under small temporal perturbations (\cref{thm:2}). Second, distributing small and opposed phase offsets across heads provably reduces local temporal variation of the modulation kernel while preserving each head’s spectrum magnitude, which realizes a controlled moving average at aggregation (\cref{thm:3}). Third, under Nyquist-valid sampling, per-head recovered spectra are invariant to temporal shifts, so PAS alters how modulation is sampled and aggregated rather than what each head encodes (\cref{thm:4}).

Evidence spans both scope and cost. Under matched token budgets, we evaluate on nine benchmarks that include action recognition and general video LLM suites, and observe consistent gains with negligible computational overhead. Parameter sweeps show wide performance plateaus and robust behavior across offset choices; sampling rate ablations align with theory, with the largest gains under sparse sampling and diminishing benefits as sampling densifies. We provide usage guidance and defaults, including simple two group settings that work well in practice.

Our contributions are threefold:
\begin{enumerate}
  \item We identify and formalize a core limitation of applying text-oriented RoPE to the time axis of video. A Fourier analysis shows that the IFT derived time domain modulation is intrinsically rippled at the frame scale, which can suppress attention to key frames in low gain intervals and induce sharp changes between adjacent frames.
  \item We introduce Phase Aggregated Smoothing (PAS), a plug-and-play, training-free mechanism that assigns opposed phase offsets to different heads while shifting only the query stream. Each head preserves the original frequency spectrum magnitude, and standard head aggregation yields a smoother effective modulation over time.
  \item We provide theory informed scope and empirically validated usage recommendations that link expected gains to sampling rate and to the temporal spectrum used by M-RoPE, covering both Nyquist valid and sub-Nyquist settings, together with negligible compute overhead under matched token budgets.
\end{enumerate}

%% file: sec/2_related_work.tex
\section{Related Work}

\noindent\textbf{Videos as Input to Transformers.}
Early Vision Transformers (e.g., TimeSformer~\cite{bertasius2021spacetimeattentionneedvideo}, ViViT~\cite{arnab2021vivitvideovisiontransformer}, Video Swin~\cite{liu2021videoswintransformer}, and MViT~\cite{fan2021multiscalevisiontransformers}) structure space-time attention and tokenization to control computational cost~\cite{dosovitskiy2021imageworth16x16words}.
These designs motivate low frame rates and frame merging in practice, which in turn increases temporal phase sensitivity.
Our approach is orthogonal to these architectures and aims to improve robustness to sampling offsets without changing token budgets or backbones.

\noindent\textbf{Positional Encodings for Video and Multimodal Settings.}
Rotary positional embeddings (RoPE) inject position via complex-plane rotations and are widely used in language and vision Transformers~\cite{su2023roformerenhancedtransformerrotary}. Extending RoPE to images and videos, including M-RoPE, raises questions about axis coupling and frequency allocation. This motivated vision-specific analyses of 2D and 3D rotary designs, as well as multimodal formulations that interleave temporal and spatial channels in a unified scheme~\cite{heo2024rotarypositionembeddingvision, wei2025videoropemakesgoodvideo, wang2024qwen2vlenhancingvisionlanguagemodels}. Alternatives based on relative positions or lightweight gating offer different trade-offs between expressiveness and cost~\cite{Hao_2024}. Our work focuses on absolute temporal encoding and studies temporal robustness from a Fourier perspective.

\noindent\textbf{Temporal Robustness: Frame Rates, Sampling Offsets, and Test-time Strategies.}
A substantial body of work shows that action recognition accuracy is sensitive to frame rate and sampling choices. Lowering FPS or shifting sampling windows can change which motion cues are aggregated per token and may flip predictions on borderline cases~\cite{schiappa2023largescalerobustnessanalysisvideo}. Benchmarks that emphasize short-term object interactions further expose this fragility under sparse sampling~\cite{10.1145/2425836.2425909}. Architectural responses include dual-rate pathways that separate slow semantic context from fast motion dynamics~\cite{Feichtenhofer_2019_ICCV}. At inference time, multi-pass strategies improve stability by averaging predictions across resampled offsets or by temporally consistent adaptation, but they increase latency and complexity~\cite{yi2023temporalcoherenttesttimeoptimization}. In contrast, we target single-pass coverage of multiple temporal phases by distributing small phase offsets across attention heads.

\noindent\textbf{Training-free baselines.}
Complementary to the efficiency and positional-encoding lines above, several training-free Video-LLM strategies expand temporal coverage at inference. SlowFast-LLaVA constructs dual pathways, where the Slow branch uses low FPS and high spatial detail and the Fast branch uses high FPS with aggressive downsampling. The two streams are concatenated to stabilize decisions under sparse sampling and motion aliasing~\cite{xu2024slowfastllavastrongtrainingfreebaseline}. TS-LLaVA (``thumbnail and sampling'') mixes a compact thumbnail that preserves global spatial cues with a lightly sampled temporal stream, providing broader time coverage without changing weights~\cite{qu2024tsllavaconstructingvisualtokens}. They can be layered on top of the same backbone as ours and directly address failure modes induced by low FPS and frame merging, which makes them ideal comparators to our approach.

%% file: sec/4_methodology.tex
\section{Methodology}

\subsection{Overview}
\label{subsec:overview}

\noindent\textbf{Problem.}
Multimodal RoPE encodes time by assigning a fixed set of temporal frequency lines, each contributing a periodic dependence of attention on the relative lag between video tokens. The aggregate time-domain modulation from these lines is intrinsically rippled at the bin scale, which is a property of the spectrum itself rather than a side effect of sparse sampling or frame merging. As a result, very small changes in lag can produce large fluctuations in attention (\cref{fig:intro-attn-instability}); at certain lags, the modulation suppresses attention, while at others it amplifies it (\cref{fig:intro-figure}). This non-smooth dependence on time makes attention overly sensitive to minor timing perturbations and weakens phase robustness, shifting control of the logits away from content similarity toward incidental timing.

\noindent\textbf{Our idea: smooth the IFT waveform at inference time.}
We introduce {Phase Aggregated Smoothing} (PAS), a plug-in that applies small phase shifts across attention heads while keeping the tokenization and parameters unchanged.
Each head observes the same merged-frame tokens but samples the phase modulation at slightly different temporal offsets. Individual head spectra remain unchanged, yet their aggregation attenuates the high-frequency ripples of the IFT waveform.
The effective result is a smoother temporal modulation that stabilizes attention against small time shifts, improving phase robustness without altering the underlying positional semantics or increasing compute.

\noindent\textbf{Roadmap.}
We briefly review RoPE on the temporal axis and state a phase modulation approximation that underpins all subsequent analyses. Using this view, we show that smoother IFT waveforms yield logits that vary more gently with relative phase, and that PAS provably reduces local variation by aggregating slightly shifted observations across head groups. We then specify the condition under which RoPE is left intact. In practice, each head’s recovered spectrum is unchanged, so PAS alters only how modulation is sampled and aggregated, not what is encoded.

\subsection{Preliminaries}
\label{sec:preliminaries}

\noindent \textbf{Background.}
RoPE encodes positions by applying rotations to the query and key in $m$ two dimensional subspaces indexed by a frequency set $\Omega=\{\omega_i\}_{i=0}^{m-1}$. Each subspace corresponds to the standard RoPE coordinate pair $(2i,\,2i{+}1)$, which can be identified with a complex scalar. Let
$z_i := q_{2i} + \mathrm{j}\, q_{2i+1}$ and $w_i := k_{2i} + \mathrm{j}\, k_{2i+1}$ denote the complexified content on pair $i$ before rotation. At position $s$, RoPE multiplies the $i$th components by the phase $e^{\mathrm{j}\omega_i s}$, that is, it applies a planar rotation in the corresponding two dimensional subspace. With $C_i := z_i w_i^{*}$, which is content dependent and independent of displacement, the RoPE-rotated attention logit depends only on the {relative displacement} $\Delta$:
\begin{equation}
\label{eq:rope-unified}
\big\langle \tilde{\mathbf{q}}, \tilde{\mathbf{k}} \big\rangle(\Delta)
= \mathrm{Re}\!\left[\sum_{i=0}^{m-1} C_i\, e^{\mathrm{j}\omega_i \Delta}\right].
\end{equation}
Here $\Delta=p-p'$ for 1D discrete indices, and $\Delta=\alpha(t-t')$ for continuous time with a fixed scale $\alpha>0$. Intuitively, RoPE compares positions through phase differences along the finite set of spectral lines $\Omega$.

\Cref{eq:rope-unified} shows that the logit is a content inner product modulated by a structured factor determined by $\Omega$ only. We formalize this as a scalar phase modulation and use it to analyze how PAS smooths temporal behavior.

\begin{theorem}[Phase modulation approximation] 
\label{thm:1}
Let $\Omega=\{\omega_i\}_{i=0}^{m-1}$ be the RoPE lines and define the IFT kernel
$
m(\Delta)\;:=\;\frac{1}{m}\sum_{i=0}^{m-1} e^{\mathrm{j}\omega_i \Delta}.
$
Assume (i) a large number of lines $m$, (ii) near-uniform spectral energy across lines (the content-dependent coefficients $\{C_i\}$ have similar magnitudes), and (iii) a fixed time scale $\alpha$ with $\Delta=\alpha(t-t')$.
Then the RoPE rotated attention satisfies
\begin{equation}
\big\langle \tilde{\mathbf{q}}, \tilde{\mathbf{k}} \big\rangle(\Delta)
\;\approx\;
\big\langle \mathbf{q}, \mathbf{k} \big\rangle \cdot \mathrm{Re}\!\big\{\,m(\Delta)\,\big\}.
\label{eq:phase-mod-approx}
\end{equation}
\end{theorem}
That is, the unrotated content dot product is multiplied by a scalar phase modulation factor given by the IFT of the RoPE spectrum evaluated at the relative displacement $\Delta$, which separates content from relative position.

With many lines and no strong content-frequency coupling, the sum in \cref{eq:rope-unified} concentrates around its average pattern, yielding the scalar modulation in \cref{eq:phase-mod-approx}.
Because $m(\Delta)$ is inherently rippled at the bin scale, tiny lag changes can cause large logit swings (temporal jitter).
PAS introduces small, opposed {temporal offsets} across head groups and aggregates them. This samples $m(\Delta)$ at nearby displacements and averages the results, which smooths the effective modulation over time while keeping each head’s recovered spectrum unchanged under Nyquist-valid sampling.
A full set of assumptions and finite-$m$ deviation bounds are provided in the supplementary material.

\subsection{Theory}
\label{subsec:theory}

We begin by making the qualitative temporal jitter precise. Along the temporal axis, multimodal RoPE induces a time-lag modulation: the attention logit at lag $\Delta t$ follows the inverse Fourier view introduced earlier, so small shifts in $\Delta t$ probe different points on the same IFT waveform. When this waveform is locally rippled, a tiny time shift can swing the modulation drastically, which indicates that attention is sensitive to phase. Our first result formalizes the opposite regime: the smoother the IFT waveform, the steadier the attention with respect to relative phase. This statement is the mathematical hinge for PAS: if we can effectively smooth the IFT waveform without altering per-head spectra, we can suppress jitter while preserving positional semantics.

\begin{theorem}[Smooth IFT $\Rightarrow$ Phase-Stable Attention] 
\label{thm:2}
Let $m(\Delta t)$ denote the IFT kernel of the RoPE frequency set (the phase modulation factor), and let $A(\Delta t)$ denote the RoPE-rotated attention logit at time lag $\Delta t$. Under the phase modulation approximation $A(\Delta t)\approx \langle \mathbf{q},\mathbf{k}\rangle\,\mathrm{Re}\{m(\Delta t)\}$, define the global slope of the kernel by $L_m:=\sup_{\tau\in\mathbb{R}}\big|\partial_\tau\,\mathrm{Re}\{m(\tau)\}\big|$. Then for any $\Delta t$ and any small increment $\delta t$,
\begin{equation}
\label{eq:stability-bound}
\big|A(\Delta t+\delta t)-A(\Delta t)\big|
\;\le\;
\big|\langle \mathbf{q},\mathbf{k}\rangle\big|\; L_m\; |\delta t|.
\end{equation}
\end{theorem}
In words, the change of the attention logit across a time shift is Lipschitz in $|\delta t|$, with a constant proportional to the kernel’s maximal local slope. A smoother IFT waveform with a smaller $L_m$ implies tighter stability of attention with respect to relative phase differences. A complete proof and an exact spectral version are provided in the supplementary material.

The IFT kernel $m(\cdot)$ plays the role of a temporal window multiplying the content dot product. If $m$ varies slowly with lag, neighboring lags receive nearly the same modulation, so the content, instead of the phase, dominates the attention decision. Conversely, when $m$ oscillates rapidly, adjacent lags are multiplied by substantially different factors, amplifying tiny timing perturbations into large logit swings. This is the failure mode that PAS targets by reducing the local slope of the effective modulation at inference time.

We now turn intuition into mechanism. \Cref{thm:2} explains why the smoothness of the modulation kernel controls the sensitivity of RoPE-rotated logits to small temporal shifts. If temporal instability stems from sampling a rippled IFT waveform at slightly different lags, then a natural remedy is to average nearby lags. PAS implements this at inference: different heads apply small, opposed phase shifts to the query, so each head examines the same merged frame tokens at slightly shifted lags of the modulation waveform, and the model then aggregates across heads. \Cref{thm:3} formalizes that this multiphase aggregation reduces the local temporal variation of the modulation, thereby suppressing jitter while preserving the spectrum of each head and providing a direct bridge to the next result.

\begin{theorem}[Multi-Phase Averaging Smooths the IFT Waveform]
\label{thm:3}
Let $m(\Delta t)$ denote the IFT kernel of the RoPE spectrum (the phase modulation factor). Consider $H$ heads with small temporal shifts $\{\delta_h\}_{h=1}^{H}$ and nonnegative aggregation weights $\{a_h\}_{h=1}^{H}$ satisfying $\sum_{h} a_h=1$.
Define the {effective modulation} after PAS aggregation by
\begin{equation}
\label{eq:m-eff}
m_{\mathrm{eff}}(\Delta t)\;:=\;\sum_{h=1}^{H} a_h\, m(\Delta t+\delta_h).
\end{equation}
For any fixed increment $\varepsilon\in\mathbb{R}$, define the mean-square local variation functional
\begin{equation}
\label{eq:var-func}
\mathcal{V}_{\varepsilon}(f)\;:=\;\limsup_{T\to\infty}\,\frac{1}{T}\int_{0}^{T}\big(f(\tau+\varepsilon)-f(\tau)\big)^{2}\,d\tau.
\end{equation}
Then
\begin{equation}
\label{eq:var-ineq}
\mathcal{V}_{\varepsilon}\!\big(m_{\mathrm{eff}}\big)\;\le\;\mathcal{V}_{\varepsilon}\!\big(m\big),
\end{equation}
with strict inequality whenever the shifts are not all identical (modulo the period induced by $\varepsilon$) and at least two weights are positive.
\end{theorem}
Equivalently in the frequency domain, the line spectrum of $m_{\mathrm{eff}}$ at angular frequency $\omega$ is multiplied by $K(\omega):=\sum_{h}a_h e^{\mathrm{j}\omega\alpha\delta_h}$, so $|K(\omega)|\le 1$ with strict attenuation $|K(\omega)|<1$ for nonzero frequencies whenever the phases $\{\omega\alpha\delta_h\}$ are dispersed. Consequently, the multi-phase aggregation damps high-frequency ripples of the IFT waveform, yielding a smoother effective modulation. A complete proof and additional variants are provided in the supplementary material.

\Cref{eq:m-eff} is a short-range temporal average of the IFT kernel. Each head samples $m(\cdot)$ at a slightly shifted lag and the model aggregates these samples. Averaging nearby points reduces local differences: the mean of small shifts changes less from $\Delta t$ to $\Delta t+\varepsilon$ than any single sample, so the characteristic wiggles of $m$ are muted.
In the frequency view, the shift-and-average corresponds to multiplying the discrete line spectrum by $K(\omega)$, a convex combination of unit phasors. Its magnitude is bounded by $1$ and strictly smaller when phases spread out, which is exactly the condition created by PAS’s opposed, small phase offsets across heads.
Through ~\cref{thm:2} and ~\cref{thm:3}, the chain is complete: {multi-phase averaging} $\Rightarrow$ {smoother effective $m$} $\Rightarrow$ {more phase-stable attention}.

Having shown why smoothing the modulation helps and how PAS achieves such smoothing by multi-phase averaging, we now address a safety question: does PAS change what RoPE encodes?
Fourier analysis tells us that time shifts do not alter a signal’s spectrum. PAS operates by per-head time shifts of the IFT waveform before aggregation.
Thus, if the temporal sampling of the video respects Nyquist, each head should preserve the RoPE spectrum; only phases rotate.
This is the guarantee that lets PAS smooth {how} modulation is sampled without rewriting {what} spectrum is present.

\begin{theorem}[Per-Head {Recovered} Spectrum is Invariant under Temporal Offsets]
\label{thm:4}
Let $m(\Delta t)$ be the IFT kernel induced by the temporal RoPE line set.
Fix a sampling period $\Delta>0$ and an $N$-point observation with a {fixed} deterministic window $w[n]$ ($0\!\le n\!<\!N$).
Define the observed sequences
$
x[n]:=w[n]\;m(n\Delta),\,x_\delta[n]:=w[n]\;m(n\Delta+\delta),
$
and their $N$-point DFTs $X[k]$ and $X_\delta[k]$.
Assume $m$ is Nyquist-band-limited, i.e., its continuous-time spectrum is supported in $|\omega|<\pi/\Delta$, so that fractional delays are implementable as all-pass phase ramps.
Then there exists a bin phase $\theta_k(\delta)$ (depending only on $\delta$, $w$, and the DFT bin $k$) such that for all $k\in\{0,\dots,N-1\}$,
\begin{equation}
\label{eq:head-spectrum-windowed}
X_\delta[k] \;=\; e^{\mathrm{j}\theta_k(\delta)}\, X[k]
\quad\Longrightarrow\quad
\bigl|X_\delta[k]\bigr| \;=\; \bigl|X[k]\bigr|.
\end{equation}
\end{theorem}
\noindent\textbf{Remark on finite sampling length.}
Because the observation length $N\Delta$ is almost invariably shorter than the period of the lowest frequency component of RoPE, the spectrum cannot be exactly recovered from $x[n]$ by any $N$-point DFT. \Cref{eq:head-spectrum-windowed} therefore states an invariance of the {recovered DFT spectrum} per head, not of the unobservable, infinite-length spectrum.
When Nyquist is violated, fractional delays cease to be all-pass, and the conclusion need not hold.

\!\!\!\!\!\!\!\!\! \textbf{Remark on temporal order across heads.}
Let the temporal delays be $\{\delta_h\}$ with $\delta_h\in[\delta_{\min},\delta_{\max}]$ and assume $\delta_{\max}-\delta_{\min}\le\Delta$ (in particular, $\delta_{\max}-\delta_{\min}\le 1$ when $\Delta$ is measured in one-bin units).
For grid points $t_i=i\Delta$ we then have:
for any $j\ge i+1$ and any heads $h,h'$, $t_i+\delta_h \le t_j+\delta_{h'}$, so adjacent or non-adjacent bins from different heads never swap their discrete order.

Consequently, the frame/bin order is preserved across heads: each head induces a shift without permuting the discrete temporal order.

Within a fixed, finite observation window, applying a small temporal offset to one head corresponds to an all-pass fractional delay whose frequency response has unit modulus.
Hence the $N$-point DFT of that head is modified only by a per-bin phase factor, leaving magnitudes unchanged.
PAS assigns opposed tiny offsets across heads, so each head’s {recovered} spectrum (DFT magnitudes) is identical to the baseline. Smoothing emerges only {after} standard multi-head aggregation, which averages phase-shifted observations in time and attenuates nonzero lines in frequency via the aggregation kernel.

\noindent \textbf{Summary.}
We diagnose temporal instability as sampling a rippled IFT modulation induced by temporal M-RoPE, so tiny lag changes can rotate the modulation away from content and dominate the logit. We then establish three theorems that link this diagnosis to a remedy. First, if the IFT waveform is smooth at the bin scale, the RoPE-rotated logit varies at most linearly with the lag perturbation, which yields stability to small timing errors. Second, averaging several small and opposed phase shifts of the same kernel acts like a moving average in time and reduces local oscillation, thereby tightening the Lipschitz control on logit variation. Third, although a finite window cannot fully recover the RoPE spectrum, the per-head recovered spectrum computed on the observed sequence is exactly invariant to such phase shifts, so the mechanism changes how the modulation is sampled and aggregated, not what each head encodes in the measured spectrum.

\subsection{Implementation}
\label{subsec:implementation}
We now present the core algorithm and separate high-level logic from engineering details needed for robust integration.

\begin{algorithm}[t]
\caption{Phase Aggregated Smoothing (single-pass inference)}
\label{alg:mph}
\begin{algorithmic}[1]
\Require Tokenized sequence $X$ with video/text masks; head count $K$; offsets $\{\delta_h\}_{h=1}^K$; weights $\{w_h\}$ with $\sum_h w_h=1$; unit {bin} or {frame} (depending on whether frames are merged).
\State Identify the video-token span (right-aligned masking) and restrict subsequent steps to video tokens.
\For{each attention layer}
  \State Compute standard $Q,K,V$ after the base positional encoding.
  \For{head $h=1$ to $K$}
     \State Map $\delta_h$ to the temporal half-dimensions (convert units if needed).
     \State Apply the phase operator only to the query stream: $Q_h \leftarrow \phaseop{\delta_h}(Q_h)$.
  \EndFor
  \State Compute attention with modified $(Q,K,V)$ and continue the normal multi-head aggregation.
\EndFor
\State \Return model output with head-wise phase diversification.
\end{algorithmic}
\end{algorithm}

\noindent\textbf{Core algorithm (\cref{alg:mph}).}
(i) $\phaseop{\delta}$ acts {only} on temporal half-dimensions, leaving spatial encoding intact. (ii) Use symmetric $\{\delta_h\}$ and simple weights (for example, uniform) to attenuate local ripple in the effective modulation. (iii) All heads observe the same merged-frame tokens, differing only by their phase-shifted queries.

\begin{figure}[t]
  \centering
  \includegraphics[width=\linewidth, clip, trim=1 328 455 23 ]
  {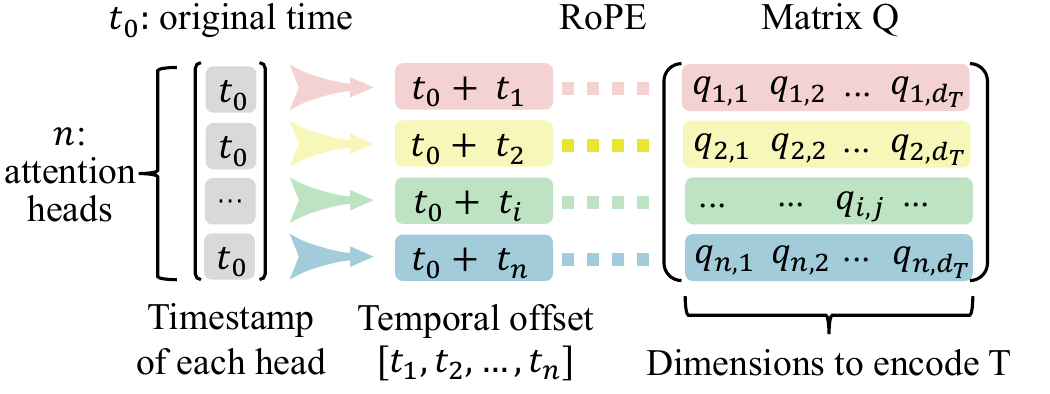}
  \caption{Implementation of PAS. For the $Q$ matrix of each vision token, we apply temporal offsets to each query head.}
   \label{fig:PAS}
\end{figure}

\noindent\textbf{Engineering details.}
(i) \emph{Hook point}: apply $\phaseop{\delta}$ \emph{after} the base positional encoding is applied to $Q$. (ii) \emph{Masking}: ensure only video tokens are affected. (iii) {MHA~\cite{vaswani2023attentionneed} / GQA~\cite{ainslie2023gqatraininggeneralizedmultiquery} compatibility}: broadcast offsets along the per-head dimensions of $Q$. (iv) \emph{Unit conversion}: keep consistency with the time definition when using {bin} or {frame} units.

\noindent\textbf{Compute overhead and throughput.}
Let $B$ be the batch size, $H$ the number of attention heads, $S$ the full sequence length, $S_v\!\le\!S$ the number of video tokens, and $d_h$ the per-head channel size. Denote by $d_t$ the number of \emph{temporal half-dimensions} affected by the phase operator (a fixed fraction $p_t\!\in\!(0,1)$ of $d_h$). The dominant per-layer attention cost scales as
\begin{equation}
    C_{\text{attn}}
    \;=\; \Theta\big(B\,H\,S^2\,d_h\big),
    \label{eq:attn-cost}
\end{equation}
while PAS adds only a per-token, per-head linear transform on the temporal half-dimensions of $Q$:
\begin{equation}
    C_{\text{PAS}}
    \;=\; \Theta\big(B\,H\,S_v\,d_t\big).
    \label{eq:mph-cost}
\end{equation}
This yields the overhead ratio
\begin{equation}
    \frac{C_{\text{PAS}}}{C_{\text{attn}}}
    \;\le\;
    \frac{S_v\,d_t}{S^2\,d_h}
    \;=\;
    \frac{p_t\,S_v}{S^2},
    \label{eq:overhead-ratio}
\end{equation}
which is typically negligible per layer since $p_t$ is a small constant and $S$ is in the hundreds to thousands.

\noindent\textbf{Inference throughput.}
On an NVIDIA A100 80\,GB and under matched sequences and settings, the original backbone achieves $(77.2\pm3.1)\times10^3$ tokens/s throughput, while PAS achieves $(76.8\pm4.0)\times10^3$ tokens/s.
The two are statistically indistinguishable within variance, indicating that enabling PAS has a negligible impact on inference speed.

%% file: sec/5_experiment.tex
\section{Experiment}

\begin{table*}[t]
\centering
\begin{threeparttable}
\caption{Training-free, matched-token results including stacked variants. PAS applies per-head offsets $[0,0.5]$ in bin units. Rows above the horizontal rule list the backbone and single-method baselines; rows below show PAS stacked with other methods. Color coding: blue marks the best score within the upper block (above the rule) for each metric column, and green marks the best score over the entire column.}

\label{tab:main-results}
\scriptsize
\setlength{\tabcolsep}{3pt}
\renewcommand{\arraystretch}{1.06}
\rowcolors{2}{gray!15}{white}
\begin{tabular}{@{}l cc cc cc cc c c c c c @{}}
\toprule
& \multicolumn{2}{c}{20BN-Jester} & \multicolumn{2}{c}{\makecell{Something-\\Something V2}} & \multicolumn{2}{c}{Kinetics-700} & \multicolumn{2}{c}{\makecell{Breakfast\\Actions}} & \multicolumn{1}{c}{MVBench} & \multicolumn{1}{c}{TempCompass} & \multicolumn{1}{c}{PerceptionTest} & \multicolumn{1}{c}{EgoSchema} & \multicolumn{1}{c}{MMBench-Video}  \\
Method & Acc & Macro-F1 & Acc & Macro-F1 & Acc & Macro-F1 & Acc & Macro-F1 & Acc & Overall & Overall & Acc & Mean Score (0–3) \\
\cmidrule(lr){2-3}\cmidrule(lr){4-5}\cmidrule(lr){6-7}\cmidrule(lr){8-9}\cmidrule(l){10-10}\cmidrule(l){11-11}\cmidrule(l){12-12}\cmidrule(l){13-13}\cmidrule(l){14-14}
Default Setting & 16.0 & 9.9 & 14.8 & 13.7 & 44.9 & 41.0 & 44.5 & 41.6 & 67.2 & 71.5 & 67.8 & 63.5 & 1.71 \\
SlowFast-LLaVA~\cite{xu2024slowfastllavastrongtrainingfreebaseline} & 14.9 & 9.3 & \textcolor{blue}{16.8} & \textcolor{blue}{14.7} & 45.1 & 42.3 & 44.2 & 41.7 & 69.2 & \textcolor{blue}{73.5} & 68.0 & 63.6 & 1.76 \\
TS-LLaVA~\cite{qu2024tsllavaconstructingvisualtokens} & 15.4 & 8.9 & 15.1 & 12.6 & 44.7 & 40.8 & 43.6 & 41.0 & 67.8 & 71.4 & \textcolor{blue}{69.3} & \textcolor{blue}{65.8} & 1.70 \\
PAS (Ours) & \textcolor{blue}{18.3} & \textcolor{blue}{12.6} & 16.3 & 14.4 & \textcolor{blue}{48.2} & \textcolor{blue}{45.1} & \textcolor{blue}{45.3} & \textcolor{blue}{42.4} & \textcolor{blue}{69.5} & 73.3 & 68.9 & 63.9 & \textcolor{blue}{1.78} \\
\midrule
SlowFast + PAS & \textcolor{ForestGreen}{19.6} & \textcolor{ForestGreen}{12.9} & \textcolor{ForestGreen}{17.4} & \textcolor{ForestGreen}{15.5} & \textcolor{ForestGreen}{49.8} & \textcolor{ForestGreen}{46.0} & \textcolor{ForestGreen}{45.6} & 42.3 & \textcolor{ForestGreen}{71.0} & \textcolor{ForestGreen}{73.9} & 69.1 & 64.1 & \textcolor{ForestGreen}{1.81} \\
TS-LLaVA + PAS & 19.3 & 12.8 & 16.5 & 14.6 & 49.1 & 45.9 & 45.3 & \textcolor{ForestGreen}{42.4} & 70.1 & 71.7 & \textcolor{ForestGreen}{69.4} & \textcolor{ForestGreen}{66.1} & 1.76 \\
\bottomrule
\end{tabular}
\end{threeparttable}
\end{table*}
\subsection{Datasets and Evaluation}
\label{sec:experiments-datasets}

We evaluate nine benchmarks in two categories and summarize them in \Cref{tab:datasets}. The table lists the task type, a brief challenge descriptor, and whether our effective temporal sampling meets the Nyquist condition for the temporal RoPE lines.

\noindent\textbf{Action recognition.}
{20BN-Jester}~\cite{9022297} emphasizes directional and ordering cues. {Something-Something V2}~\cite{Goyal_2017_ICCV} focuses on object manipulations, and {Kinetics-700}~\cite{smaira2020shortnotekinetics7002020human} covers diverse motions. For these three datasets, the sampling satisfies Nyquist, so PAS is tested in a phase-stable regime. {Breakfast Actions}~\cite{Kuehne_2014_CVPR} contains long activities; we use sub-sampled clips that are below Nyquist to assess behavior under aliasing.

\noindent\textbf{General video-LLM suites.}
{MVBench}~\cite{li2024mvbenchcomprehensivemultimodalvideo} targets quasi-static perception, {TempCompass}~\cite{liu2024tempcompassvideollmsreally} isolates temporal perception, {PerceptionTest}~\cite{pătrăucean2023perceptiontestdiagnosticbenchmark} probes multimodal reasoning, {EgoSchema}~\cite{mangalam2023egoschemadiagnosticbenchmarklongform} evaluates long-horizon understanding, and {MMBench-Video}~\cite{fang2024mmbenchvideolongformmultishotbenchmark} measures multi-shot comprehension.

\begin{table}[t]
\centering
\scriptsize
\setlength{\tabcolsep}{6pt}
\renewcommand{\arraystretch}{1.05}
\caption{Benchmarks used in our evaluation. The Nyquist column indicates whether the effective temporal sampling meets the Nyquist condition for the temporal RoPE line set.}
\label{tab:datasets}
\begin{tabular}{@{}lccc@{}}
\toprule
Dataset & Type & Key challenge & Nyquist? \\
\midrule
20BN-Jester~\cite{9022297} & Action & Directional cues & $\checkmark$ \\
Something-Something V2~\cite{Goyal_2017_ICCV} & Action & Object manipulation & $\checkmark$ \\
Kinetics-700~\cite{smaira2020shortnotekinetics7002020human} & Action & Diverse motions & $\checkmark$ \\
Breakfast Actions~\cite{Kuehne_2014_CVPR} & Action & Long, sub-sampled & $\times$ \\
MVBench~\cite{li2024mvbenchcomprehensivemultimodalvideo} & General & Motion analysis & $\checkmark$ \\
TempCompass~\cite{liu2024tempcompassvideollmsreally} & General & Temporal dependency & $\checkmark$ \\
PerceptionTest~\cite{pătrăucean2023perceptiontestdiagnosticbenchmark} & General & Multimodal reasoning & $\checkmark$ \\
EgoSchema~\cite{mangalam2023egoschemadiagnosticbenchmarklongform} & General & Long-horizon & $\checkmark$ \\
MMBench-Video~\cite{fang2024mmbenchvideolongformmultishotbenchmark} & General & Multi-shot & $\checkmark$ \\
\bottomrule
\end{tabular}
\end{table}

\subsection{Main Results}
\label{subsec:main-results}

\noindent\textbf{Setup.} We compare six inference strategies: the original backbone \texttt{Qwen2.5-VL-7B-Instruct}, two public training-free baselines (SlowFast-LLaVA and TS-LLaVA), our PAS, and two stacked variants that apply PAS on top of SlowFast-LLaVA and TS-LLaVA. All methods use identical decoding and matched video-token budgets per clip. Baseline inputs follow public recipes while keeping total tokens comparable. For PAS, we partition attention heads into $K{=}2$ phase groups and apply per-head offsets $[0,\,0.5]$ in bin units to the backbone query stream, acting only on video tokens after the base positional encoding. For stacked variants, we keep each baseline’s input recipe unchanged and apply the same offsets; token budgets remain matched.

\noindent\textbf{Results.}
\Cref{tab:main-results} summarizes the training-free, matched-token comparison. On the phase-sensitive action benchmarks, PAS consistently improves over the backbone and attains the best single-model results on 20BN-Jester and Kinetics-700, while stacked variants set the strongest overall results on SSv2 and remain competitive on Breakfast. On Breakfast (sub-Nyquist), all methods show modest changes consistent with aliasing-limited gains, and stacking yields only small additional improvements, matching the scope of \cref{thm:4}. On the diagnostic and general video-LLM suites, PAS improves over the backbone across MVBench, TempCompass, PerceptionTest, EgoSchema, and MMBench-Video, and stacking further raises the top line, achieving the best numbers in multiple columns. Overall, $K{=}2$ with offsets $[0,\,0.5]$ remains a strong default, and PAS layers cleanly on multi-rate or thumbnail pipelines without training, input changes, or added tokens.

\subsection{Parameter Insensitivity of PAS}
\noindent\textbf{Goal.}
We substantiate two claims: (i) on a fixed dataset, PAS requires no hyper-parameter fine-tuning and exhibits wide performance plateaus, and (ii) there exists a common hyper-parameter range that transfers across datasets. We evaluate the four datasets from \cref{sec:experiments-datasets} under identical token budgets and preprocessing.

\noindent\textbf{Controls and scope.}
Unless noted, all runs share the same uniform frame sampling and per-bin frame merging as in \cref{subsec:main-results}, the same decoding settings, and the same video-token budget. PAS is applied only to video tokens after the base positional encoding and only to the query stream. Plots report {classification accuracy} (Acc).

\noindent\textbf{Preliminaries.}
A {bin} is the unit produced by the sampler and merger (a contiguous bucket of frames merged into one video token along time).
Let $H_q$ denote the number of query heads in the backbone (GQA or MHA).
We introduce a {phase group} parameterization with $K\!\ge\!2$ groups that evenly partition the $H_q$ query heads into disjoint sets $\{\mathcal{G}_g\}_{g=0}^{K-1}$; all heads within the same group share a single temporal offset $\phi_g$ (expressed in {bin units}, where $\phi{=}1.0$ equals a shift of one bin).
This grouping reduces tunable hyperparameters from $H_q$ per-head offsets to only $K$ group offsets and makes the setting readily {transferable} across models with different head counts. Unless otherwise stated, in this subsection, we fix $K{=}2$ and use $(\phi_0,\phi_1)=(0,\Delta)$.

\begin{figure}[t]
  \centering
  \begin{subfigure}{.495\linewidth}
    \includegraphics[width=\linewidth]{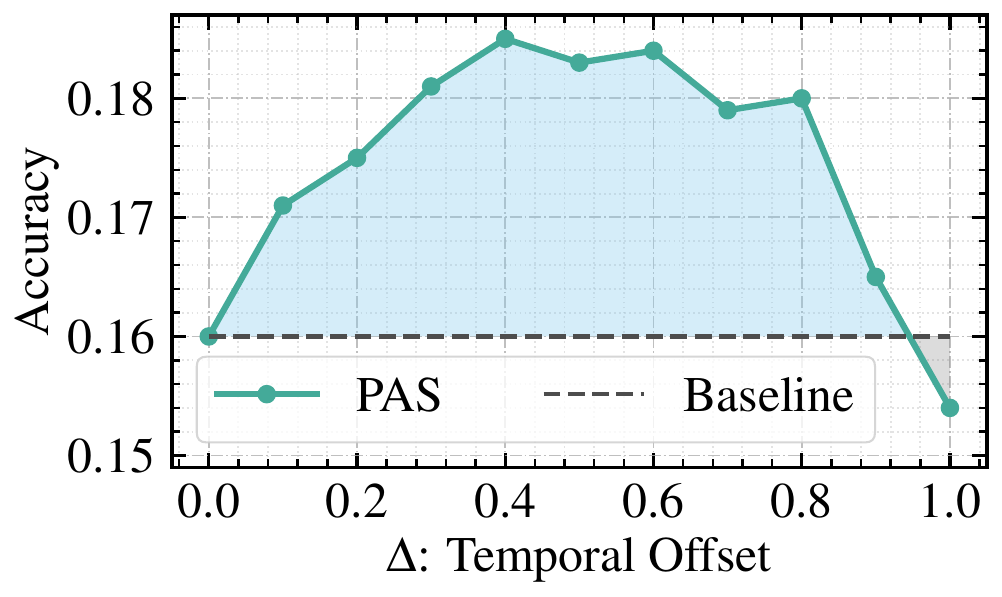}
    \caption{20BN-Jester}
    \label{fig:k2-delta-jester}
  \end{subfigure}\hfill
  \begin{subfigure}{.495\linewidth}
    \includegraphics[width=\linewidth]{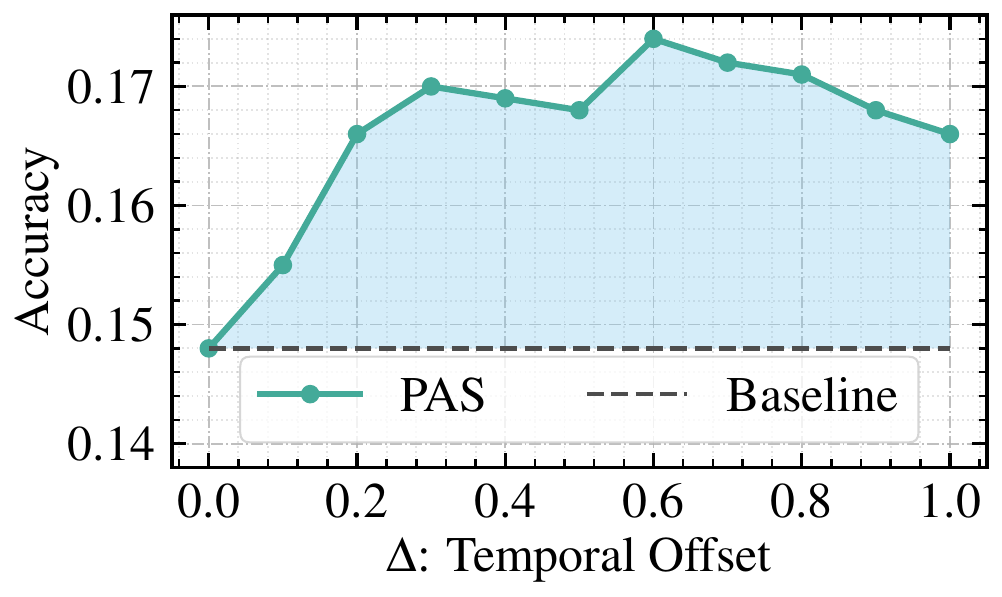}
    \caption{Something-Something V2}
    \label{fig:k2-delta-ssv2}
  \end{subfigure}
  
  \vspace{0.5em}
  
  \begin{subfigure}{.495\linewidth}
    \includegraphics[width=\linewidth]{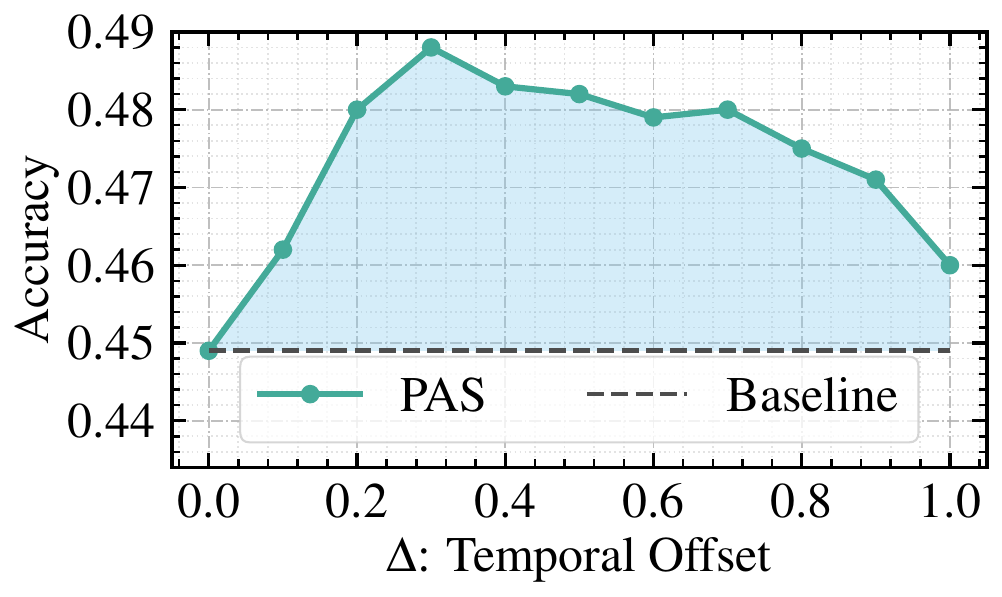}
    \caption{Kinetics-700}
    \label{fig:k2-delta-k700}
  \end{subfigure}\hfill
  \begin{subfigure}{.495\linewidth}
    \includegraphics[width=\linewidth]{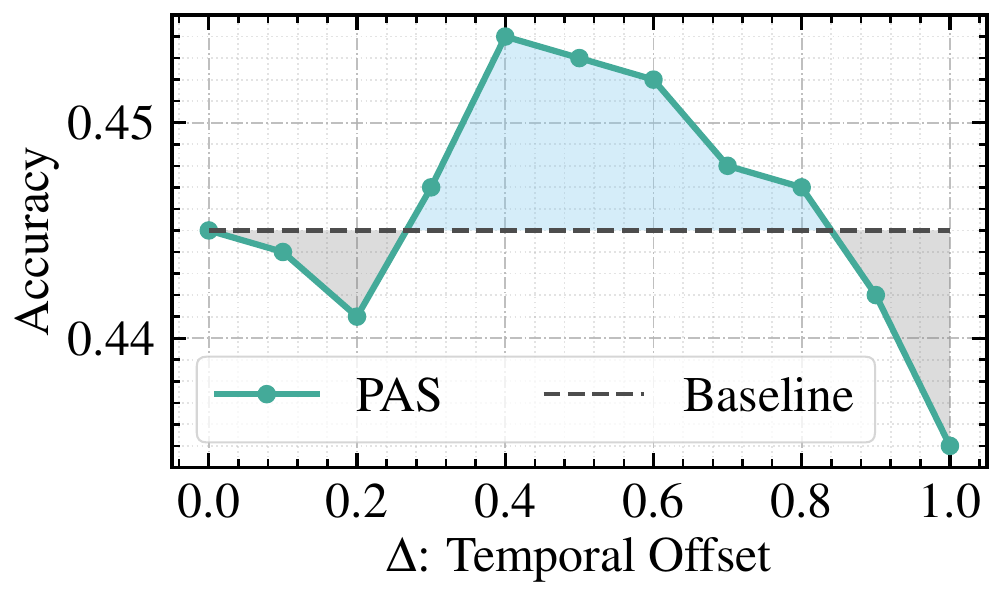}
    \caption{Breakfast Actions}
    \label{fig:k2-delta-breakfast}
  \end{subfigure}
  \vspace{-0.3em}
  \caption{Offset sweep with fixed $K{=}2$. Offsets $\Delta$ are applied to only one of the two groups of query heads in bin units. We report accuracy as a function of $\Delta$.}
  \label{fig:k2-delta-scan}
\end{figure}

\noindent\textbf{$\Delta$ scan at fixed $K{=}2$.}
We sweep a single magnitude $\Delta \in \{0.0,\,0.1,\,0.2,\,\dots,\,1.0\}$ (step $0.1$). $\Delta{=}0$ degenerates to the single-phase baseline. Values near $\Delta{\approx}1.0$ approach the bin-scale aliasing boundary. For each dataset, we plot {Acc} versus $\Delta$; see \cref{fig:k2-delta-scan}. From \cref{thm:3}, increasing $\Delta$ spreads the phases and increases attenuation on nonzero lines through $|K(\omega)|$, up to the point where large shifts begin to probe different portions of $m(\cdot)$ that may no longer yield additional smoothing.

\noindent\textbf{Findings.}
Across the three motion-rich datasets (20BN-Jester, SSv2, Kinetics-700), we observe {stable and significant} gains for $\Delta$ in the range $0.3$ to $0.8$. For Breakfast Actions (undersampled), the improvement window is narrower and the absolute gains are smaller. These patterns align with our theory: when Nyquist is satisfied, per-head spectra are preserved (\cref{thm:4}) and multi-phase averaging primarily smooths temporal modulation (\cref{thm:3}) which tightens the Lipschitz bound on logit changes (\cref{thm:2}). Under undersampling, aliasing limits the effective benefit. A practical default is $\Delta{\approx}0.5$, which lies well inside the cross-dataset plateau.

\subsection{Ablation: Impact of Sampling Rate on PAS}
\label{subsec:ablation-sampling}

\noindent\textbf{Goal.}
To test whether PAS’s gains indeed come from smoothing the time-domain modulation implied by RoPE’s spectrum, we vary the {sampling ratio} $r\in[0,1]$ (sampled frames divided by total frames per video). As $r$ increases, adjacent samples are closer in time, the IFT waveform is probed more densely, and the phase modulation difference between neighboring bins shrinks. We therefore expect PAS’s advantage to diminish as $r$ increases, matching the prediction of \cref{thm:2} that smaller local slopes leave less room for additional smoothing.

\noindent\textbf{Setup.}
All settings match \cref{subsec:main-results} except that $r$ is the independent variable. For each $r$, we compare the original backbone and PAS under identical preprocessing and decoding, with the same token budget at each sampling ratio $r$. PAS uses $K{=}2$ head groups with per-head offsets $(0,\Delta)$ fixed at $\Delta{=}0.5$ (bin units), applied only to the query stream after the base positional encoding.

\begin{figure}[t]
  \centering
  \begin{subfigure}{.495\linewidth}
    \includegraphics[width=\linewidth]{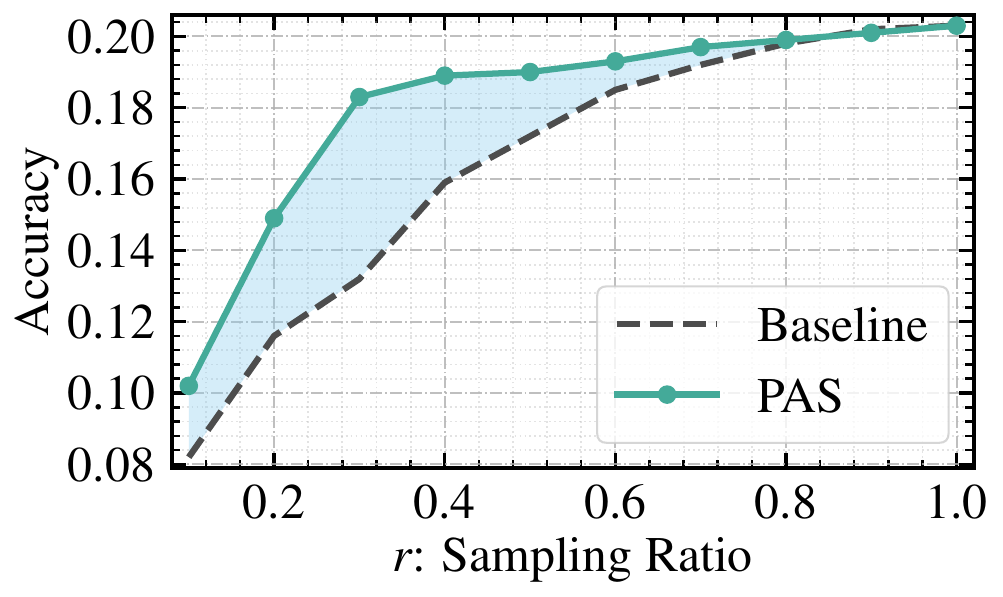}
    \caption{20BN-Jester}
    \label{fig:ablation-r-jester}
  \end{subfigure}\hfill
  \begin{subfigure}{.495\linewidth}
    \includegraphics[width=\linewidth]{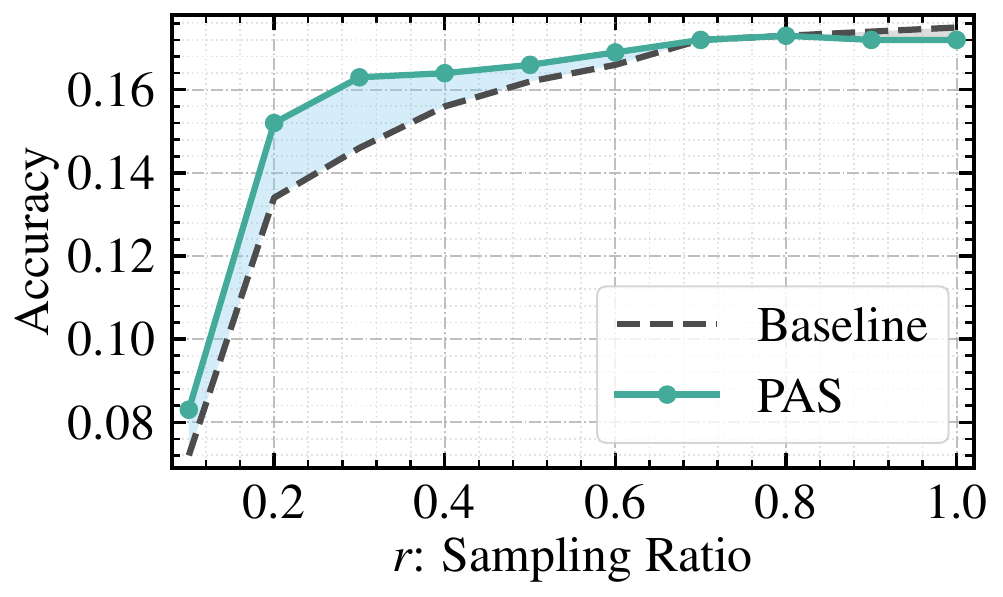}
    \caption{Something-Something V2}
    \label{fig:ablation-r-ssv2}
  \end{subfigure}
  
  \vspace{0.5em}
  
  \begin{subfigure}{.495\linewidth}
    \includegraphics[width=\linewidth]{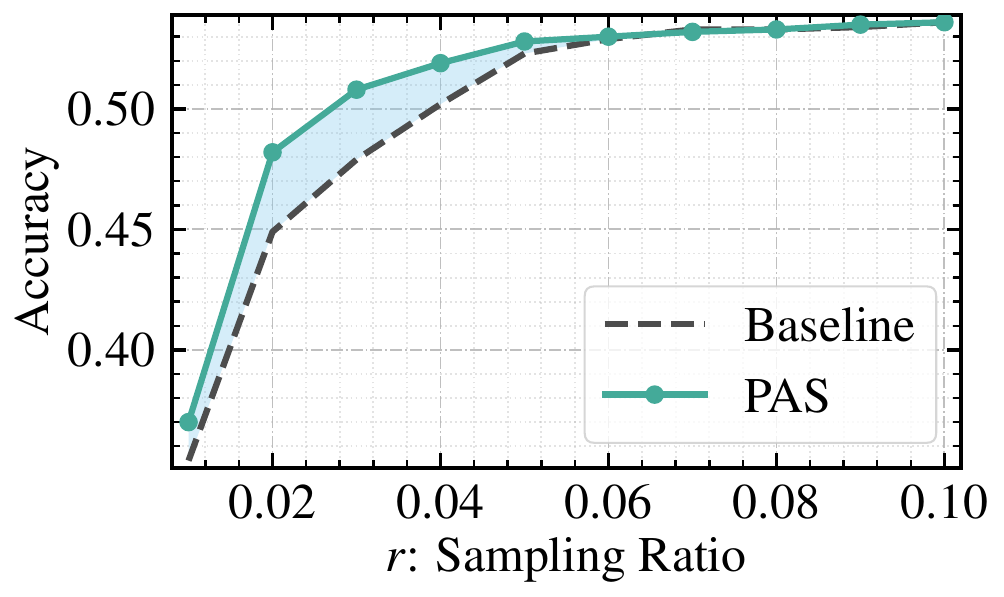}
    \caption{Kinetics-700}
    \label{fig:ablation-r-k700}
  \end{subfigure}\hfill
  \begin{subfigure}{.495\linewidth}
    \includegraphics[width=\linewidth]{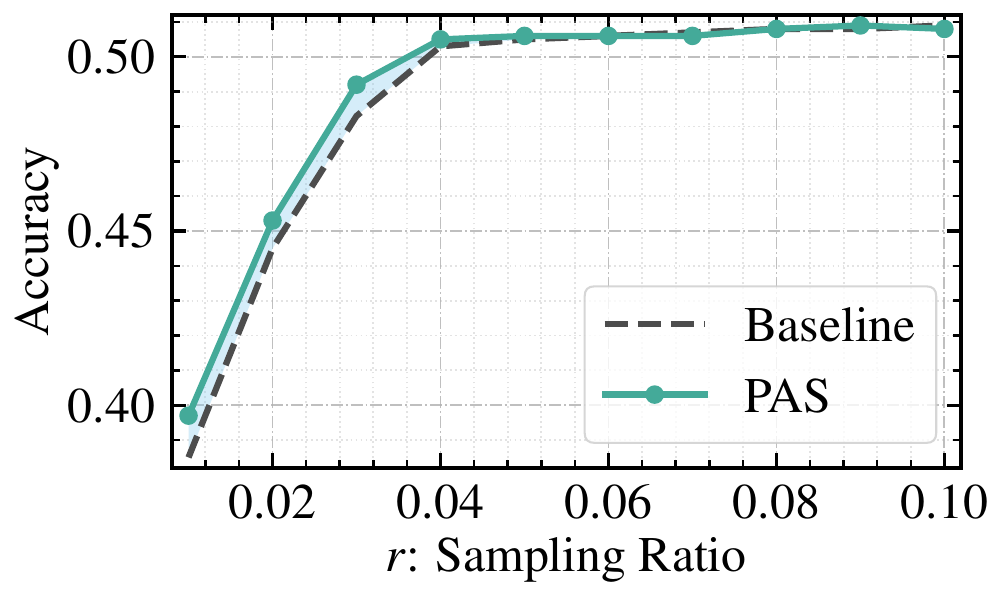}
    \caption{Breakfast Actions}
    \label{fig:ablation-r-breakfast}
  \end{subfigure}
  \vspace{-0.3em}
  \caption{Sampling ratio ablation with fixed $K{=}2$, $\Delta{=}0.5$. Classification accuracy as a function of the sampling ratio $r$.}
  \label{fig:ablation-r}
\end{figure}

\noindent\textbf{Findings.}
Across 20BN-Jester, SSv2, and Kinetics-700, PAS yields larger gains at lower sampling ratios and gradually converges to the original backbone as $r$ increases. When $r$ is high, the two curves are statistically indistinguishable. On Breakfast Actions (undersampled relative to the RoPE band), the improvement window is narrower and the absolute gains are smaller. These observations are consistent with our theory: denser sampling intrinsically smooths the effective probing of the IFT kernel (smaller empirical local slopes, as in \cref{thm:2}), leaving less headroom for inference-time smoothing. Under sparse sampling, PAS compensates by averaging nearby phase observations across heads (\cref{thm:3}) while preserving per-head spectra whenever Nyquist holds (\cref{thm:4}).

%% file: sec/6_conclusion.tex
\section{Conclusion}

We set out from a simple diagnosis: temporal M-RoPE makes attention depend on the IFT of a line spectrum, and under sparse sampling this waveform ripples so that tiny time shifts cause large logit swings. Our answer is {Phase Aggregated Smoothing} (PAS), a training-free, inference-time plug-in that applies phase offsets across heads so that aggregation averages nearby lags. This smooths the modulation introduced by the temporal encoding while leaving each head’s recovered spectrum unchanged whenever Nyquist holds. The theory closes the loop: smoother IFT leads to phase-stable attention (\cref{thm:2}); multi-phase averaging smooths the waveform and attenuates nonzero lines (\cref{thm:3}); per-head temporal shifts preserve spectral magnitudes (\cref{thm:4}). A minimal implementation that touches only $Q$ on temporal dimensions yields negligible overhead. Under matched token budgets, experiments validate consistent gains, broad offset plateaus, and the predicted trend that benefits taper as sampling densifies. In short, PAS changes how modulation is sampled and aggregated, not what is encoded, offering a practical path to robust Video LLMs in low-FPS, frame-merged regimes.

%% file: sec/7_acknowledgements.tex
\section*{Acknowledgements}

The work is supported in part by the U.S. National Science Foundation (NSF) under Grant CRII 2451683, an NVIDIA Academic Grants Program, a U.S. Bank Academic Research Award, the University of California, Merced, a UC Merced Faculty Research Award, and the Institute of Information \& Communications Technology Planning \& Evaluation (IITP) grant funded by the Korean Government (MSIT) (No. RS-2024-00457882, National AI Research Lab Project). The views and conclusions are those of the authors and do not necessarily reflect the official policy or position of the U.S. Government or other funding agencies.